\documentclass{article}
\usepackage{ijcai21}

\usepackage{times}

\usepackage{soul}
\usepackage{url}
\usepackage[hidelinks]{hyperref}
\usepackage[utf8]{inputenc}
\usepackage[small]{caption}
\usepackage{graphicx}
\usepackage{amsmath}
\usepackage{booktabs}
\usepackage{multirow}
\usepackage{blindtext}
\usepackage{tabularx}
\usepackage[table,xcdraw]{xcolor}
\usepackage{array}
\usepackage{url}
\usepackage{color}
\usepackage{booktabs}
\usepackage{multirow}
\usepackage{graphicx}
\usepackage{xspace}
\usepackage{soul}
\usepackage{fancyhdr}
\usepackage{csquotes}

\newcommand{\latinphrase}[1]{\textit{#1}}

\newcommand{\ie}{\latinphrase{i.e.}\xspace}

\newcommand{\etc}{\latinphrase{etc.}\xspace}

\title{Densely Deformable Efficient Salient Object Detection Network}

\begin{document}

\author{
Tanveer Hussain$^1$\and
Saeed Anwar$^{2,3,4}$\and
Amin Ullah$^1$\and
Khan Muhammad$^1$\And \\
Sung Wook Baik$^1$\footnote{Corresponding author: Sung Wook Baik}\\
\affiliations
$^1$Sejong University, Seoul, South Korea, 
$^2$The Australian National University, Australia,\\ 
$^3$Data61-CSIRO, Australia, 
$^4$University of Technology Sydney, Australia.
\emails
\{tanveer445, aminullah, khan.muhammad\}@ieee.org,
saeed.anwar@data61.csiro.au,
sbaik @sejong.ac.kr}

\maketitle
\begin{abstract}

Salient Object Detection (SOD) domain using RGB-D data has lately emerged with some current models' adequately precise results. However, they have restrained generalization abilities and intensive computational complexity. In this paper, inspired by the best background$/$foreground separation abilities of deformable convolutions, we employ them in our Densely Deformable Network (DDNet) to achieve efficient SOD. The salient regions from densely deformable convolutions are further refined using transposed convolutions to optimally generate the saliency maps. Quantitative and qualitative evaluations using the recent SOD dataset against 22 competing techniques show our method's efficiency and effectiveness. We also offer evaluation using our own created cross-dataset, surveillance-SOD (S-SOD), to check the trained models' validity in terms of their applicability in diverse scenarios. The results indicate that the current models have limited generalization potentials, demanding further research in this direction. Our code and new dataset will be publicly available at \url{https://github.com/tanveer-hussain/EfficientSOD}

\end{abstract}

\section{Introduction}
Salient object detection (SOD) highlights important and distinctive contents from image data naturally observable by the human visual system. SOD is useful in various computer vision applications, including activity recognition \cite{ullah2020conflux}, video summarization \cite{hussain2021comprehensive}, \etc With the emergence of CNNs, researchers have achieved significantly precise SOD results, surpassing the traditional hand-crafted features-based methods \cite{cheng2014depth}. Most current methods follow a supervised learning strategy, using saliency maps with their corresponding RGB and depth images for training. We follow the same strategy but consider only RGB images to generate saliency maps.

The SOD methods can be categorized based on input data and fusion strategy. Firstly, as the RGB input utilizes a single CNN network without any fusion, where the input is only RGB; secondly, the RGB and depth is input separately to a two-stream network with late fusion; thirdly, providing RGB and depth combinedly to the network, called as early fusion; and finally, the one that involves fusion at multiple stages in the network.

\begin{figure} 
\begin{tabular}[b]{c@{}c@{}c@{}c@{}c@{}c} 
\raisebox{1.2\normalbaselineskip}[0pt][0pt]{\rotatebox{90}{Images}}&      
\includegraphics[width=.09\textwidth]{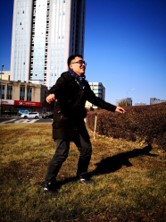}&
\includegraphics[width=.09\textwidth]{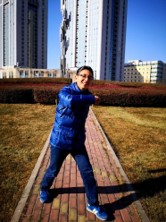}&   
\includegraphics[width=.09\textwidth]{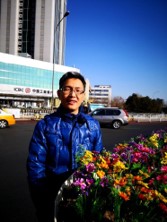}&
\includegraphics[width=.09\textwidth]{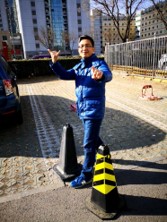}&
\includegraphics[width=.09\textwidth]{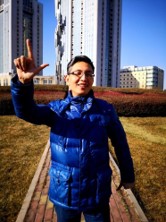}\\

\raisebox{2\normalbaselineskip}[0pt][0pt]{\rotatebox{90}{GT}}&
\includegraphics[width=.09\textwidth]{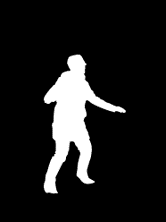}&
\includegraphics[width=.09\textwidth]{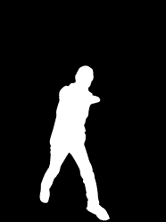}&   
\includegraphics[width=.09\textwidth]{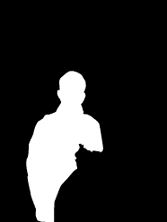}&
\includegraphics[width=.09\textwidth]{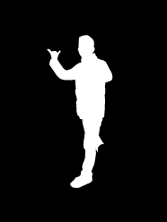}&
\includegraphics[width=.09\textwidth]{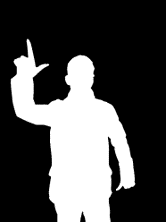}\\ 

\raisebox{1.5\normalbaselineskip}[0pt][0pt]{\rotatebox{90}{DDNet}}&
\includegraphics[width=.09\textwidth]{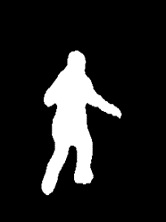}&
\includegraphics[width=.09\textwidth]{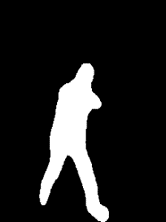}&   
\includegraphics[width=.09\textwidth]{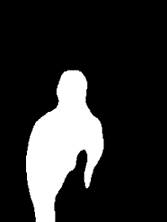}&
\includegraphics[width=.09\textwidth]{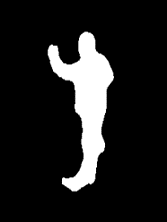}&
\includegraphics[width=.09\textwidth]{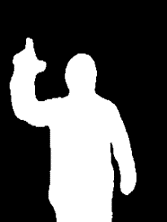}\\
& a)  & b) & c) & d) & e)\\
\end{tabular}
    \caption{Applying our DDNet over challenging test images from the SIP dataset, the last row shows proposed results. (a) The right hand of the person is half covered, correctly segmented by DDNet; (b) complex pose with half hand overlapping with the body; (c) half of the body is covered with another foreground and correctly differentiated using DDNet; (d) right foot covered behind another foreground object, accurately distinct-ed by DDNet; (e) even the fingers of the person are correctly segmented by DDNet, as given in the GT.}
    \label{fig:sampleresults}
\end{figure}

The SOD methods based on RGB~\cite{deng2018r3net} have limited results in challenging scenarios when objects have a similar texture and homogeneous properties with complex scattered backgrounds. Similarly, the RGB-D or two-steam networks are computationally expensive \cite{zhang2020uc}, inefficient, and have limited adaptability in real-world applications. Moreover, the RGB-D methods for SOD show comparatively poor results when depth images have noise or heavy occlusions. This paper achieves efficient and SOTA SOD without depth data (see Figure \ref{fig:sampleresults}), thereby significantly fewer training parameters. DDNet employs densely deformable convolutions to model geometrical transformations in training data effectively, enabling generalization potentials. The transpose convolutions with bi-linear interpolation for image smoothing and a basic image processing algorithm as post-processing generates refined saliency maps, covering the aforementioned limitations in the existing literature.

\begin{figure} 
\begin{tabular}[b]{c@{}c@{}c@{}c@{}c@{}c} 
\raisebox{0.75\normalbaselineskip}[0pt][0pt]{\rotatebox{90}{Images}}&      
\includegraphics[width=.09\textwidth]{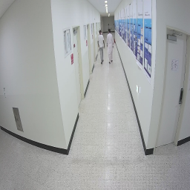}&
\includegraphics[width=.09\textwidth]{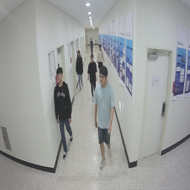}&   
\includegraphics[width=.09\textwidth]{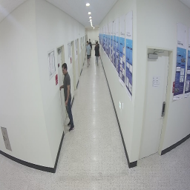}&
\includegraphics[width=.09\textwidth]{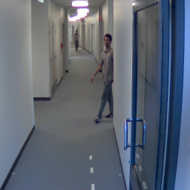}&
\includegraphics[width=.09\textwidth]{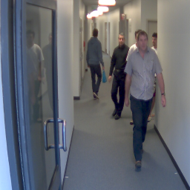}\\

\raisebox{0.9\normalbaselineskip}[0pt][0pt]{\rotatebox{90}{Depths}}&
\includegraphics[width=.09\textwidth]{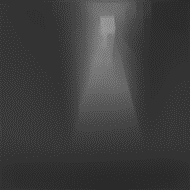}&
\includegraphics[width=.09\textwidth]{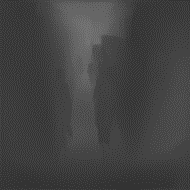}&   
\includegraphics[width=.09\textwidth]{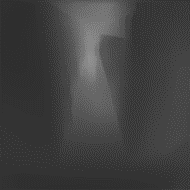}&
\includegraphics[width=.09\textwidth]{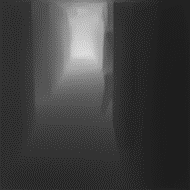}&
\includegraphics[width=.09\textwidth]{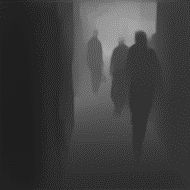}\\

\raisebox{1.5\normalbaselineskip}[0pt][0pt]{\rotatebox{90}{GT}}&
\includegraphics[width=.09\textwidth]{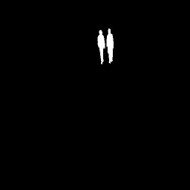}&
\includegraphics[width=.09\textwidth]{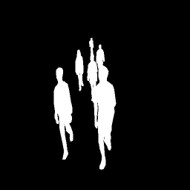}&   
\includegraphics[width=.09\textwidth]{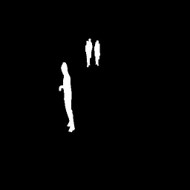}&
\includegraphics[width=.09\textwidth]{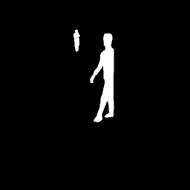}&
\includegraphics[width=.09\textwidth]{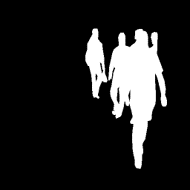}\\
& a)  & b) & c) & d) & e)\\
\end{tabular}
    \caption{Sample images from our S-SOD dataset with a diverse set of challenges. (a) objects very far from the camera with likely foreground and background; (b) multiple objects with occlusion and varying distances from the camera; (c) side-wise object and multiple objects far from the camera; (d) far and side-wise objects with varying brightness scenario; (e) multiple objects with varied poses, directions, appearance ratio, and high-level occlusion. It should be noted that our method does not use depth images; we have only provided the depths for illustration purposes.}
    \label{fig:S-SODdatasetsamples}
\end{figure}

We present a new cross-dataset, S-SOD (samples in Figure \ref{fig:S-SODdatasetsamples}), having complex images from diverse indoor surveillance scenarios to test the potentials of our SOD models, posing a lower generalization capability; hence, limited performance on open datasets. Typically, the deep SOD models learn millions of parameters to identify, discover, and locate patterns in images. The evaluation process indicates how well the SOD models perform on a certain test set. However, traditionally, this set is disjoint from the same training dataset and usually does not have comparably different samples to evaluate the model's robustness in challenging scenarios. The standard evaluation may be suitable for some background-changing applications \cite{fan2020rethinking}, but not in real-world implementation such as surveillance scenarios, where a little noise can alter the saliency maps.
In this article, we claim the following contributions:
\begin{itemize}

\item We propose a novel architecture employing densely deformable convolutions to capture and extract the salient objects' regions and pose comparatively better generalization abilities.

\item To validate the generalization abilities of SOD models, we create a small-scale dataset by collecting the most challenging images with varying brightness and contrast, background and foreground colors overlap, among many others.

\item The proposed model results show that it obtains prominent salient regions with better performance and several ablation studies with varied computation and accuracy.  Our final model achieves the best SOTA results than existing competing algorithms.

\end{itemize}

\section{Related Works}

Recently, many researchers have focused on fused modalities such as RGB-D data and fusion strategies, \ie, input fusion, early fusion, and late fusion strategies. For instance, \cite{fu2020siamese} introduced a siamese network for RGB-D data with two modules, including joint learning and densely cooperative fusion, where the first module generates robust saliency features and the second one finds the complementary features. Similarly, ICNet \cite{li2020icnet} employed an information conversion network that comprises concatenation operations and correlation layers for effective salient object detection. 

Similarly, \cite{fan2020bbs} fused RGB and depth features using the bifurcated backbone strategy to extract multi-level features, depth cues from the RGB, and depth-enhanced module, respectively. Recently, a multi-modal fusion network \cite{chen2019multi} employed two networks for SOD by utilizing dilated convolutional layers and 1$\times$1 kernels for global contextual reasoning. The first network takes RGB or depth data to predict dense area in the image, and the second network applies a late fusion strategy to combine RGB-D clues for SOD. In RGB-D based methods, one of the major challenges is cross-modal fusion. For this purpose, \cite{liu2020cross} introduced adaptive gated fusion with a two-steams Generative Adversarial network that simultaneously takes RGB and depth images utilizing depth-wise separable residual convolutions to capture the side-output feature in the RGB stream, which is later added with the depth stream decoder network to compensate the non-clarity of the depth images. The discriminator of their network fuses the stream and generates the optimal saliency maps. Working with cross-modal fusion, \cite{zhang2020select} proposed compensation-aware loss to investigate cross-modal relationships and effective fusion of their features.

\begin{figure*}[!h]
    \centering
    \includegraphics[width=\textwidth]{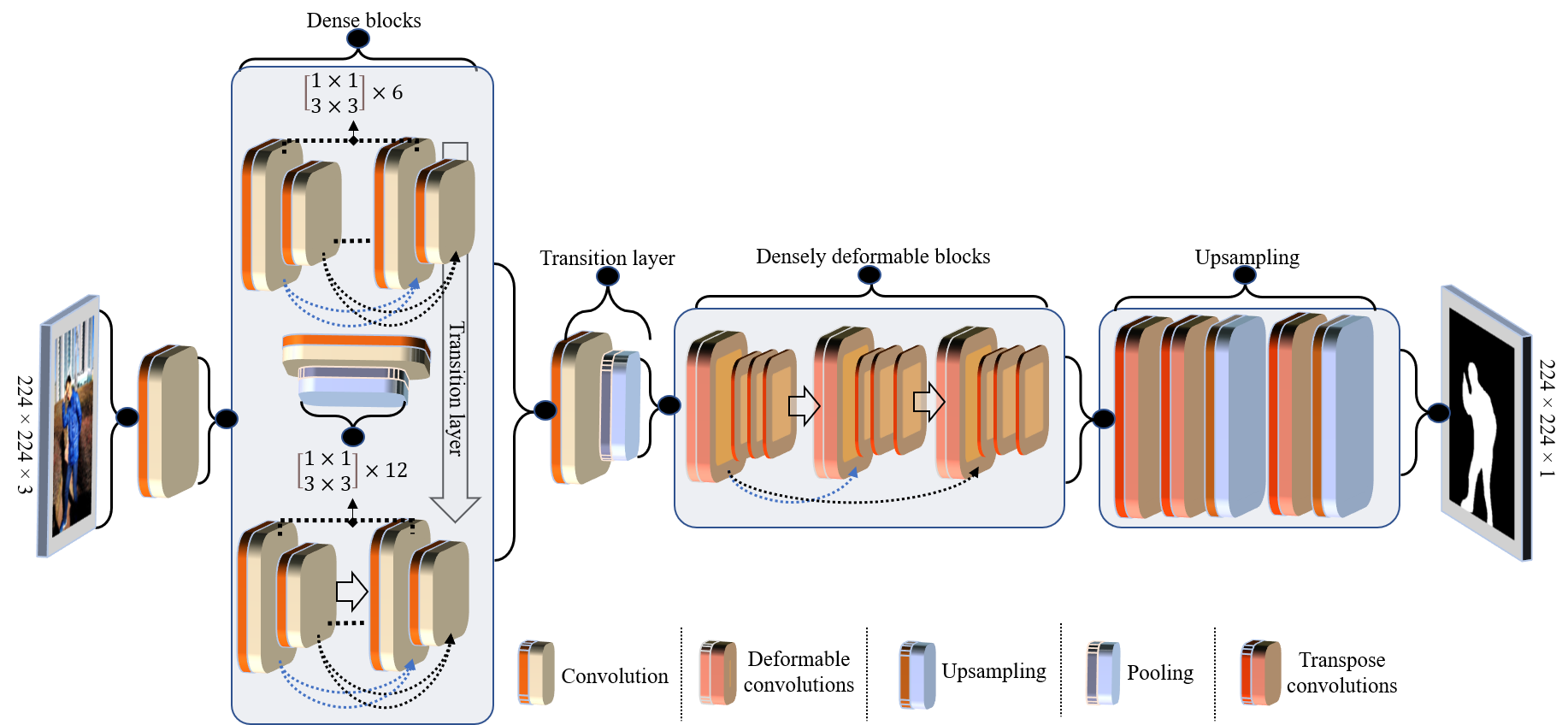}
    \caption{The proposed DDNet for efficient saliency detection that inputs only RGB image and generates its corresponding saliency map with a single color channel.}
    \label{fig:DDNet}
\end{figure*}

The techniques mentioned earlier have sophisticated results for the benchmark datasets, where training and testing data belong to the same dataset. Although traditional CNNs have shown promising results for many challenging image and video analytic domains, but some key limitations are associated with them, such as modeling geometric transformations (varied scales, poses, viewpoint, \etc) and using huge data (extensive data augmentation in some cases) for training. The existing SOD methods do not function well to model unknown transformations but show the best performance only when testing data of the same dataset is used for testing. Unlike existing methods, we employ a set of densely deformable convolutions for SOD, which has comparatively better potential to learn geometric transformations and better pattern representation abilities than traditional convolutions. Our model's details, \emph{Densely Deformable Network, abbreviated as, DDNet} are given in Figure \ref{fig:DDNet}.

\section{DDNet: Our Saliency Detection Model}
\label{DDNet}
Here we explain our proposed saliency detection model's working that uses three main blocks to generate optimal saliency. Firstly, two dense convolution blocks represent low-level features of the input RGB images. Then we propose densely connected deformable convolutions to learn effective features of salient regions and their corresponding boundaries. Finally, we employ transpose convolution and upsampling to generate the resultant saliency image. Figure~\ref{fig:DDNet} is a schematic diagram of our proposed DDNet.

\subsection{Learning Distinctive Edge and Shape Features}
In order to get the initial low-level features, such as edges, we test various convolution invariants such as dilated and grouped convolutions with pooling inserted between these layers, but their output features are not sufficient for finer details extraction. On the contrary, the densely connected convolutions proved effective in extracting high-level details. Two densely populated sequential blocks are employed in our network containing six convolution layers of kernel size 1$\times$1 and 3$\times$3, having max-pooling and batch normalization between them. We employ a transition layer before the next dense block with single 1$\times$1 convolutional layer and average pooling with a stride of two to reduce the output size. Next is the final layer of the first main block, which has densely connected convolutions with the same size as the previous one but different in numbers, \ie, a total of 12 convolutional layers for each 1$\times$1 and 3$\times$3 kernel. Our network's initial main block is inspired by the DenseNet  \cite{huang2017densely}, which has better classification results than residual networks.

The dense blocks in our network have a significant role in representing the image in terms of features. Moreover, to extract representative features, we experimented with standard convolutions and forwarded to deformable convolutions; the results were not encouraging, as documented in Table \ref{tab:ablationstudy} for various blocks of VGG16. Therefore, in our final architecture, we merely used two dense blocks. Furthermore, increasing the number of blocks may generate better results; however, at the cost of the model's efficiency and overfitting.

\subsection{Deformable Convolutions}
The building modules of CNNs do not work perfectly well to model geometric transformations inside images, as they have fixed geometric structures. Deformable convolutions and deformable RoI pooling are introduced to replace the traditional convolutions and several pooling invariants. They have better abilities to adapt geometrical variations inside an image containing objects, enabling them to generate fine saliency maps, particularly when they receive a preprocessed image from the earlier module's dense block in DDNet.

At first, we apply three deformable convolutions without any dense connection between the output of the first and the third layer's input, which results in coarse saliency maps with some roughness at the edges. Next, we employ the densely connected deformable blocks, having three interconnected convolutions in a sequence, which achieves the most significant and dominant results against SOTA, as demonstrated in Table~\ref{tab:resultscomparison}. Thus the final proposed DDNet contains densely connected three deformable convolution layers. Since our target is an efficient SOD, we did not stack more deformable convolutions, though it will be a more generic representation of input images into their corresponding saliency maps. But at the same time, the network's efficiency is compromised; hence such settings can only be adopted for applications having sufficient resources. In fact, this is the primary motivation for DDNet, to not entirely depend on deformable convolutions due to their complexity compared to regular convolutions.

\subsection{Towards Optimal Salient Image Generation}
The final block in DDNet comprises a series of transpose convolutions that pleasantly generates the image sized output with a single channel, \ie, the annotated saliency map of the corresponding input image. Some of the SOTA methods directly apply upsampling and ignore the edge information obtained via image smoothing. We achieve edge smoothing by transposing convolutions or deconvolutions and applying upsampling to resize the densely deformable block's features. Transpose convolutions in DDNet are employed using bi-linear interpolation that applies a weighted average on the distance of four corresponding nearest cells, thereby significantly smoothing the output. In our DDNet, there are three transpose convolutions and two upsampling layers.

\begin{table*}[tbp!h]
\caption{Experimental results obtained over SIP dataset against SOTA methods. Best performance is shown as bold.}
\captionsetup{justification=centering}
\rowcolors{2}{gray!25}{white}
\begin{center}
\begin{tabular}[width=\columnwidth]{l||c|ccccc}

\hline
\multicolumn{1}{c||}{Models}   & Parameters&  E-Measure $\uparrow$ & S-Measure $\uparrow$ & Weighted-F $\uparrow$ & F-Measure $\uparrow$ & MAE $\downarrow$ \\ \hline \hline

\cite{cheng2014depth}        & - & 0.770 & 0.616 & - & 0.669 & 0.298 \\ 
\cite{ju2014depth}           & - & 0.614 & 0.732 & - & 0.542 & 0.172 \\ 
\cite{ren2015exploiting}     & - & 0.759 & 0.653 & - & 0.657 & 0.185 \\ 
\cite{guo2016salient}        & - & 0.592 & 0.628 & - & 0.515 & 0.164 \\ 
\cite{cong2016saliency}      & - & 0.598 & 0.683 & - & 0.499 & 0.186 \\ 
\cite{feng2016local}         & - & 0.651 & 0.727 & - & 0.571 & 0.200 \\ 
\cite{han2017cnns}           & - & 0.705 & 0.716 & - & 0.608 & 0.139 \\ 
\cite{qu2017rgbd}            & - & 0.565 & 0.653 & - & 0.464 & 0.185 \\ 
\cite{song2017depth}         & - & 0.645 & 0.717 & - & 0.568 & 0.167 \\ 
\cite{zhao2019contrast}      & - & 0.893 & 0.850 & - & 0.821 & 0.064 \\ 
\cite{chen2019three}         & - & 0.870 & 0.835 & - & 0.803 & 0.075 \\ 
\cite{chen2019multi}         & - & 0.845 & 0.833 & - & 0.833 & 0.086 \\ 
\cite{wang2019adaptive}      & - & 0.793 & 0.720 & - & 0.702 & 0.118 \\ 
\cite{fan2020rethinking}     & - & 0.909 & 0.860 & - & 0.861 & 0.063 \\  
\cite{liu2020cross}          & - & 0.891 & 0.840 & - & 0.829 & 0.070 \\ 
\cite{fan2020bbs}            & - & 0.922 & 0.879 & - & 0.883 & 0.085 \\ 
\cite{zhang2020uc}           & - & 0.914 & 0.875 & - & 0.867 & 0.051 \\ 
\cite{zhang2020uncertainty} (w/o depth) & - & 0.927 & \textbf{0.883} & - & 0.877 & 0.045 \\
\cite{zhang2020uncertainty} (w depth)   & - & 0.927 & \textbf{0.883} & - & 0.877 & 0.045 \\
\cite{zhai2020bifurcated}    & - & 0.906 & 0.879 & - & 0.868 & 0.055 \\ 
\cite{zhang2020bilateral}    & - & 0.913 & \textbf{0.883} & - & \textbf{0.873} & 0.052 \\
\cite{fu2020jl}              & - & 0.923 & 0.879 & - & 0.885 & 0.051  \\ \hline \hline
\textbf{DDNet} (w/o DD)  & 3,334,829 & 0.917 & 0.844 & 0.762 & 0.785 & 0.050 \\ 
\textbf{DDNet} (w DD)    & 3,334,829 & \textbf{0.935} & 0.863 & \textbf{0.797} & 0.813 & \textbf{0.043}  \\ \hline \hline

\end{tabular} 
\end{center}
\label{tab:resultscomparison}
\end{table*}

\section{Experimental Results}
In this section, we explain the training and testing methods used to evaluate the performance of our DDNet and comparison with rival methods. We then introduce our newly collected dataset, S-SOD, and present various ablation studies to verify the effectiveness and efficiency of the final DDNet compared to SOTA.

\subsection{Implementation Details}
We use the PyTorch framework to implement DDNet, taking three channels input (RGB), generating a grayscale saliency map using a single NVIDIA GeForce RTX 2070 SUPER. Then, the model's dense blocks are initialized using pre-trained ImageNet weights of the same model ~\cite{huang2017densely}. Deformable convolutions are implemented with the settings mentioned in \cite{zhu2019deformable}, performing better than the standard version of the deformable convolutions. We use an ADAM optimizer with a base learning rate of $1e^{-4}$, where the training and testing images are resized to 224$\times$224 using transformers during preprocessing. We use 16 mini-batch size to train DDNet for 500 epochs with MSE loss that takes around three hours and 0.0352 to 0.0473 seconds to test an image.

\subsection{Datasets}
\textbf{Salient Person Dataset:}
Recently, Salient Person (SIP) dataset is created with 929 challenging images, and their corresponding ground truth masks and depths ~\cite{fan2020rethinking}. We follow the training and testing split\footnote{https://github.com/DengPingFan/D3NetBenchmark} by~\cite{fan2020rethinking} to compare our DDNet with 22 competing algorithms on this benchmark dataset.

\vspace{2mm}
\noindent
\textbf{A Novel Surveillance SOD Dataset:}
\label{ss:dataset}
The current salient object detection models generate near-to-human saliency maps but are limited only to the data used for training and lack generalization capability, and perform poorly due to the absence of cross-dataset validation, referred to as open data evaluation in computer vision. We build a new dataset, S-SOD, with high-resolution images and are collected to cover the most challenging scenarios for SOD methods. The challenges in S-SOD include occlusion, changes in occurrences, multiple persons' appearance, highly variable lights, and variation in distance of persons from the surveillance camera, \etc make it more generalized towards real-world data, ensuring adaptable, trained models. Most of the images are collected from the Chokepoint dataset \cite{wong_cvprw_2011}, while the rest of the challenges with varied distance and complex background and foreground are covered using our own captured RGB images.

There are 70 high-resolution images annotated using the MATLAB 2020b image segmentor tool, and the binary annotations are confirmed by two computer vision experts after being created by an unbiased user in our laboratory. The images and ground truth labels are available in 512 $ \times $ 512 size. The depth of our dataset is synthetically generated using an existing CNN-based method\footnote{https://github.com/ialhashim/DenseDepth}. We aim to eliminate depth images' dependency while generating saliency, which has high affluence in mainstream existing SOD methods. Most real-world applications deal with RGB images only, such as action recognition and persons' interactions depending on humans' saliency \ie, their localization using saliency maps helps significantly in-accurate predictions. Sample images from our S-SOD are shown in Figure~\ref{fig:S-SODdatasetsamples}.

\subsection{Metrics}
We use five metrics to compare DDNet with competing SOD algorithms. The training parameters in Table \ref{tab:resultscomparison} show the efficiency of our DDNet. The metrics are E-measure (E-M), S-measure (S-M), Weighted-F (W-F), F-measure (F-M), and Mean Absolute Error (MAE). A useful SOD model has larger E-M, S-M, W-F, F-M, and a smaller MAE. For a fair comparison, we use an existing MATLAB implementation\footnote{https://github.com/jiwei0921/Saliency-Evaluation-Toolbox/} to compute these metrics.

\subsection{Comparison with State-of-the-arts} \label{ss:comparisons}
\textbf{Quantitative comparison:} We analyze the results of the proposed DDNet against the most recent deep learning models as well as hand-crafted feature-based methods to show its dominance/superiority. DDNet achieves the lowest error rates and the highest E-M against recent competing algorithms. We surpass the best results achieved so far on the SIP dataset and determine a new SOTA SOD, as reported in Table \ref{tab:resultscomparison}. The initial rows in Table \ref{tab:resultscomparison} are hand-crafted feature-based methods for SOD, that are reported from recent works \cite{zhai2020bifurcated,zhang2020uncertainty}. DDNet achieves the lowest MAE with a 0.02 decrease from the recent UCNet \cite{zhang2020uncertainty}, and 0.02 margin against the baseline method \cite{fan2020rethinking}. DDNet also scores the highest E-M against rivals, but the S-M and F-M for DDNet are not the best when compared to some recent methods~\cite{zhang2020uncertainty,fan2020bbs}. It is worth mentioning that these methods take multiple inputs (RGB and depth) when generating saliency maps and our DDNet only depends on the RGB input, posing higher efficiency. Further, multiple inputs processing and their fusion significantly increase the computation cost, and a little noise in depth data adversely affects the saliency results.

\begin{figure*} [t]
\centering
\begin{tabular}[b]{c@{}c@{}c@{}c@{}c@{}c@{}c@{}c@{}c@{}c} 
\raisebox{2.5\normalbaselineskip}[0pt][0pt]{\rotatebox{90}{a)}}&      
\includegraphics[width=.1\textwidth]{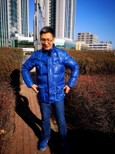} \hspace{1mm}&
\includegraphics[width=.1\textwidth]{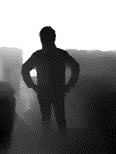} \hspace{1mm} &   
\includegraphics[width=.1\textwidth]{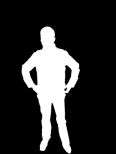} \hspace{1mm} &
\includegraphics[width=.1\textwidth]{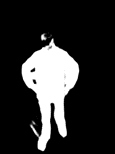} \hspace{1mm} &
\includegraphics[width=.1\textwidth]{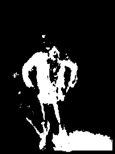} \hspace{1mm} &
\includegraphics[width=.1\textwidth]{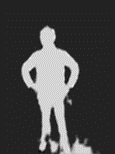} \hspace{1mm} &   
\includegraphics[width=.1\textwidth]{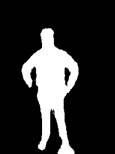} \hspace{1mm} &
\includegraphics[width=.1\textwidth]{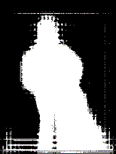} \hspace{1mm}&
\includegraphics[width=.1\textwidth]{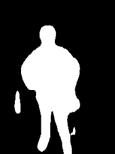}\\

\raisebox{2.5\normalbaselineskip}[0pt][0pt]{\rotatebox{90}{b)}}&
\includegraphics[width=.1\textwidth]{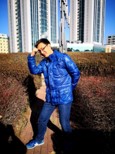} \hspace{1mm}&
\includegraphics[width=.1\textwidth]{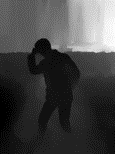} \hspace{1mm} &   
\includegraphics[width=.1\textwidth]{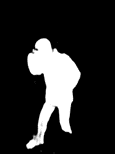} \hspace{1mm} &
\includegraphics[width=.1\textwidth]{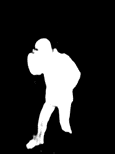} \hspace{1mm} &
\includegraphics[width=.1\textwidth]{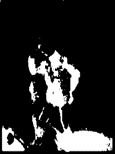} \hspace{1mm} &
\includegraphics[width=.1\textwidth]{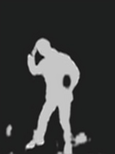} \hspace{1mm} &   
\includegraphics[width=.1\textwidth]{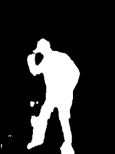} \hspace{1mm} &
\includegraphics[width=.1\textwidth]{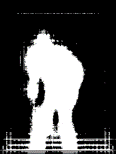} \hspace{1mm}&
\includegraphics[width=.1\textwidth]{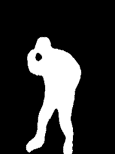}\\

\raisebox{2.5\normalbaselineskip}[0pt][0pt]{\rotatebox{90}{c)}}&
\includegraphics[width=.1\textwidth]{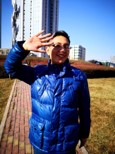} \hspace{1mm}&
\includegraphics[width=.1\textwidth]{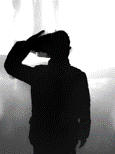} \hspace{1mm} &   
\includegraphics[width=.1\textwidth]{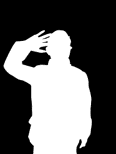} \hspace{1mm} &
\includegraphics[width=.1\textwidth]{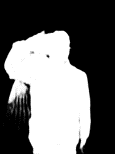} \hspace{1mm} &
\includegraphics[width=.1\textwidth]{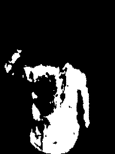} \hspace{1mm} &
\includegraphics[width=.1\textwidth]{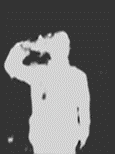} \hspace{1mm} &   
\includegraphics[width=.1\textwidth]{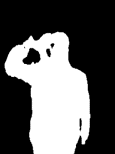} \hspace{1mm} &
\includegraphics[width=.1\textwidth]{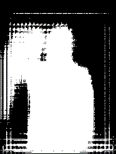} \hspace{1mm}&
\includegraphics[width=.1\textwidth]{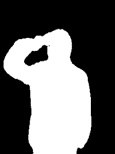}\\

\raisebox{2.5\normalbaselineskip}[0pt][0pt]{\rotatebox{90}{d)}}&
\includegraphics[width=.1\textwidth]{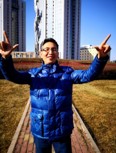} \hspace{1mm}&
\includegraphics[width=.1\textwidth]{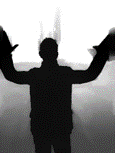} \hspace{1mm} &   
\includegraphics[width=.1\textwidth]{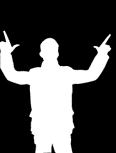} \hspace{1mm} &
\includegraphics[width=.1\textwidth]{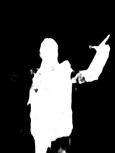} \hspace{1mm} &
\includegraphics[width=.1\textwidth]{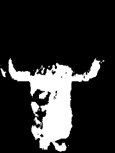} \hspace{1mm} &
\includegraphics[width=.1\textwidth]{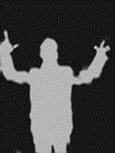} \hspace{1mm} &   
\includegraphics[width=.1\textwidth]{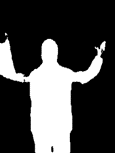} \hspace{1mm} &
\includegraphics[width=.1\textwidth]{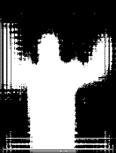} \hspace{1mm}&
\includegraphics[width=.1\textwidth]{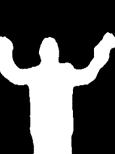}\\

\raisebox{2.5\normalbaselineskip}[0pt][0pt]{\rotatebox{90}{e)}}&
\includegraphics[width=.1\textwidth]{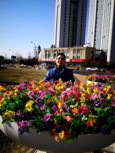} \hspace{1mm}&
\includegraphics[width=.1\textwidth]{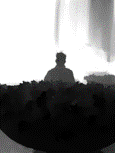} \hspace{1mm} &   
\includegraphics[width=.1\textwidth]{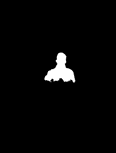} \hspace{1mm} &
\includegraphics[width=.1\textwidth]{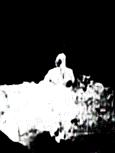} \hspace{1mm} &
\includegraphics[width=.1\textwidth]{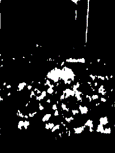} \hspace{1mm} &
\includegraphics[width=.1\textwidth]{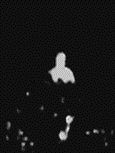} \hspace{1mm} &   
\includegraphics[width=.1\textwidth]{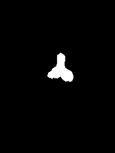} \hspace{1mm} &
\includegraphics[width=.1\textwidth]{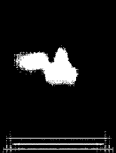} \hspace{1mm}&
\includegraphics[width=.1\textwidth]{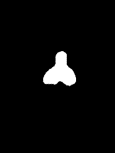}\\

& Image  & Depth & GT & UCNet & ENet &VGG16-2B & VGG16 & DDNet & DDNet \\
&  & & &  & & (SSIM) & (SSIM) & (Dilated) &  \\
\end{tabular}
    \caption{Visual comparison of saliency maps predicted by UC-Net, the proposed DDNet, and ablation models. Results improves from left to right, where DDNet outperforms others by predicting more consistent results with the ground truth (given as GT).}
    \label{fig:Visualcomparison}
    \vspace{-3mm}
\end{figure*}

\vspace{2mm}
\noindent
\textbf{Qualitative comparisons:} In Figure \ref{fig:Visualcomparison}, we visually compare DDNet with the most recent RGB-D saliency detection UC-Net \cite{zhang2020uc} over challenging images from the SIP dataset test set. We use publicly available codes and trained models of UC-Net \footnote{https://github.com/JingZhang617/UCNet} to generate these results. Figure \ref{fig:Visualcomparison} shows that UC-Net results include non-salient regions, particularly for challenging images. These images contain a complex background and occluded foregrounds with the salient object. DDNet predicts more precise saliency maps than others, indicating effectiveness and robustness. The ablation research results improve from left to right in Figure \ref{fig:Visualcomparison}, where the Dilated DDNet and VGG16 backbone with SSIM loss extract comparatively better salient regions. We will publish the results, codes, and trained models at GitHub.

\begin{table}[tbp]
\caption{Ablation study over SIP dataset using various loss functions and baseline architectures.}
\captionsetup{justification=centering}
\rowcolors{2}{gray!25}{white}
\resizebox{\columnwidth}{!}{
\begin{tabular}{l||c|cccc}\hline
        \multicolumn{1}{c||}{Models} & Loss & E-M $\uparrow$& S-M $\uparrow$& F-M $\uparrow$& MAE $\downarrow$\\ \hline \hline
        DDNet (Dilated=5) & MSE & 0.576 & 0.501 & 0.167 & 0.085 \\ 
        DDNet (Dilated=7) & MSE & 0.771 & 0.718 & 0.604 & 0.116 \\ 
        E-Net & BCE & 0.672 & 0.566 & 0.478 & 0.234 \\ 
        VGG16 (w smoothing) & MSE & 0.882 & 0.821 & 0.816 & 0.049 \\  
        VGG16-2B & MSE &  0.916 & 0.838 & 0.828 & 0.046 \\ 
        VGG16 & SSIM & 0.506 & 0.441 & 0.223 & 0.207 \\  
        VGG-2B & SSIM & 0.507 & 0.414 & 0.190 & 0.220 \\ 
        VGG16 & BCE & 0.688 & 0.581 & 0.552 & 0.189 \\  
        VGG-2B & BCE & 0.592 & 0.460 & 0.376 & 0.247 \\ \hline \hline
\end{tabular}}
\label{tab:ablationstudy}
\end{table}
\subsection{Ablation Study and Results on S-SOD}
\label{ss:ablation_S-SODresults}
\textbf{Testing Baselines:} We perform extensive experiments using various loss functions and baseline architectures to show the adequate potentials of representing an image into its corresponding features. The first block of the proposed DDNet represents low-level image features, and we perform experiments on several CNN networks such as EfficientNet that is computationally feasible, and VGG16 that functions well for geometrical variations modeling due to its complex architecture. The experimental results in Table \ref{tab:ablationstudy} show that densely connected convolutions for initial input image features representation perform better. Further, we also apply image smoothing with several parameter tuning strategies, but the saliency maps are not convincing enough. The results using efficient baselines, such as EfficientNet, are also reported in Table \ref{tab:ablationstudy}, which has the lowest number of parameters, but the saliency maps have a lower match with the ground truth in testing data.

\vspace{1mm}
\noindent
\textbf{Different Loss Functions:} The loss function used in our final experiments is MSE, but we also experimented using binary cross-entropy (BCE) and implemented structural similarity index measurement (SSIM) negation as a loss function. The ablation results in Table~\ref{tab:ablationstudy} indicate that the best results are achieved using VGG16's initial two convolutional blocks, among other options under consideration.

Finally, the proposed DDNet's results on our own created S-SOD dataset are reported in Table \ref{tab:ssodresults}, indicating lower generalization abilities of our model. Table \ref{tab:ssodresults} indicates comparatively better results for DDNet and Dilated DDNet against VGG16 and ENet, but the numbers suggest that further potential research is needed for cross dataset SOD domain.

\begin{table}[tbp]
\caption{Experimental results of DDNet over the proposed S-SOD.}\vspace{-3mm}
\rowcolors{2}{gray!25}{white}
\begin{center}
\resizebox{\columnwidth}{!}{
\begin{tabular}{l||cccc}\hline
Models & E-M $\uparrow$& S-M $\uparrow$& F-M $\uparrow$& MAE $\downarrow$\\ \hline \hline
VGG16 (w smoothing) & 0.478 & 0.305 & 0.112 & 0.402 \\
ENet  (BCE) & 0.254 & 0.061 & 0.087 & 0.900 \\ 
DDNet (Dilated) & 0.576 & \textbf{0.501} & 0.167 & \textbf{0.085} \\
DDNet  & \textbf{0.595} & 0.488 & \textbf{0.176} & 0.108 \\ \hline  \hline 
\end{tabular}}
\end{center}
\label{tab:ssodresults}
\vspace{-3mm}
\end{table}

\section{Conclusions and Future Directions}
In this paper, we proposed a densely connected and deformable convolutions-based network for salient object detection. Distinct from existing SOD methods, we achieve fine SOD using only RGB data, reducing a model's dependency and extending its applicability in real-world scenarios. The proposed DDNet is inspired by the superior features representation potentials of deformable convolutions, where we proved to achieve more accurate detecting salient objects detection results without even using depth data. Initial low-level features in the proposed DDNet are extracted using dense convolutions that are delicately refined using our dense deformable block. Learning geometrical transformation modeling using deformable blocks makes DDNet more generalized towards unseen data from a completely different environment. The final saliency is generated using transpose convolutions with bi-linear interpolation, and the results indicated the superior performance of DDNet against 22 competing SOTA methods on the SIP dataset. Even though DDNet achieved the best performance over the existing SIP dataset against rivals, its geometrical modeling and salient regions' representation is not enough to be generalized towards other domains. This aspect of SOD is alarming and attention-seeking for future research. In this direction, we introduced a very small-scaled benchmark S-SOD dataset that is collected using surveillance cameras. In the future, we aim to extend the challenges in S-SOD and increase the number of samples. Another direction is to propose an end-to-end model, inputting an RGB image, generating its corresponding depth, followed by the saliency maps generation.

\hspace{0.7mm} \textbf{Acknowledgments.} This work was supported by the National Research Foundation of Korea (NRF) Grant funded by the Korea government (MSIT) (2019R1A2B5B01070067).

\bibliographystyle{plain}
\bibliography{ijcai21}

\end{document}